\pdfoutput=1
\documentclass[11pt]{article}

\usepackage{emnlp2021}

\usepackage{times}
\usepackage{latexsym}

\usepackage[T1]{fontenc}
\usepackage[utf8]{inputenc}

\usepackage{microtype}

\usepackage{color}

\usepackage{graphicx}
\usepackage{amsmath}
\usepackage{pdfpages}
\usepackage{multirow}
\usepackage{bbm}
\usepackage{booktabs}

\newcommand{\mimlroberta}{$\text{MIML}_\text{RoBERTa}$~}

\title{Hierarchical Multi-Instance Multi-Label Learning for Detecting Propaganda Techniques}

\author{Anni Chen \and Bhuwan Dhingra \\
        Department of Computer Science
        \\ Duke University
        \\ Durham, NC, USA\\
    \texttt{anni.chen@alumni.duke.edu} \\
    \texttt{bhuwan.dhingra@duke.edu}}

\begin{document}
\maketitle
\begin{abstract}
Since the introduction of the SemEval $2020$ Task $11$ \citep{task}, several approaches have been proposed in the literature for classifying propaganda
based on the rhetorical techniques used to influence readers.
These methods, however, classify one span at a time, ignoring dependencies from the labels of other spans within the same context.
In this paper, we approach propaganda technique classification as a
Multi-Instance Multi-Label (MIML) learning problem \citep{miml} and propose a simple RoBERTa-based model \citep{roberta} for classifying all spans in an article simultaneously. Further, we note that, due to the annotation process where
annotators classified the spans by following a decision tree,
there is an inherent hierarchical relationship among the different
techniques, which existing approaches ignore. We incorporate these hierarchical label dependencies by adding an auxiliary classifier for each node in the decision tree to the training objective and ensembling the predictions from the original and auxiliary classifiers at test time. Overall, our model leads to an absolute improvement of $2.47\%$ micro-F1 over the model from the shared task winning team in a cross-validation setup and is the best performing non-ensemble model on the shared task leaderboard.

\end{abstract}
\section{Introduction}

The development of the Web and social media has amplified the scale and effectiveness of propaganda \citep{barron2019proppy}. 
Automatic propaganda detection through text analysis enables social science researchers to analyze its
spread at scale \cite{glowacki2018news, martino2020survey}. Within text analysis, two broad approaches include
the identification of rhetorical techniques used to influence readers
and document-level propagandistic article classification.
The former is more promising because propaganda techniques are easy to identify and are the very building blocks \citep{martino2020survey}. Hence, it is the focus of this paper.

% Propaganda becomes rampant in everyday news and online posts, as seen in the US presidential election and the global infodemic amid the COVID-19 pandemic \citep{martino2020survey}.

% woolley2017computational, martino2020survey

% Propaganda becomes rampant in everyday news and online posts, as seen in the Russian use of propaganda in the US presidential election \citep{ghosh2018digital}, social bots spreading propaganda in the Brazil election \citep{arnaudo2017computational}, the constant use of propaganda in Chinese governance \citep{bolsover2017computational} and the global infodemic amid the COVID-19 pandemic \citep{patwa2021fighting}.

% Further identifying the specific techniques used for influencing readers can also help 
% train readers to spot propaganda and develop targeted approaches to combat it.
% To counter the massive propaganda with speed, automatic systems are in the best position to help perform the detection task.
% However, the black-box nature of the models (TODO:\cite[]{kausar2020prosoul, barron2019proppy}) which only gives a binary classification in detecting propaganda texts presents a major challenge to their wide adoption (TODO:\cite[]{yu2021interpretable)}. Thus, it is crucial to have a model that identifies the specific propaganda technique for a propagandistic text so as to increase the interpretability and hence the trustworthiness of the automatic systems (TODO:\cite[]{yu2021interpretable, martino2020survey}). 

Recent research on fine-grained propaganda detection has been spurred by the NLP4IF-2019 Shared Task \citep{da-san-martino-etal-2019-findings} and its follow-up SemEval $2020$ Task $11$ \citep{task}.
In this paper, we focus on its Technique Classification subtask where,
given a news article text and spans identified as propagandistic, systems need to
classify each of the spans into one or more of $14$ different common propaganda techniques.

All top systems submitted to the task \citep{applicaai, aschern, hitachi, solomon} employed a 
pretrained RoBERTa-based model \citep{roberta} which was trained to classify one span at a time. 
However, labels of different spans within the same article clearly depend on each other. Thus, we approach the task within a Multi-Instance Multi-Label framework \citep{miml}, where we model
each article as an object with multiple instances (spans), each with its own labels.
This allows us to model the dependencies between different labels within the same article.
We show that this $\text{MIML}_\text{RoBERTa}$ observes a $1.98\%$ micro-F1 improvement over the replicated ApplicaAI system (referred to as baseline) \citep{applicaai}.  

% In the presence of expected interdependencies, modeling across several span fragments jointly is preferred over learning a single span independently, in a similar vein as the Conditional Random Field for the Name Entity Recognition Task where strong grammatical rules are governing the tagging process(TODO:CRF).
% \bd{Not sure if comparison to CRFs is appropriate here. }

\begin{figure*}
\centering
\small
\begin{center}
\includegraphics[width=0.85\textwidth]{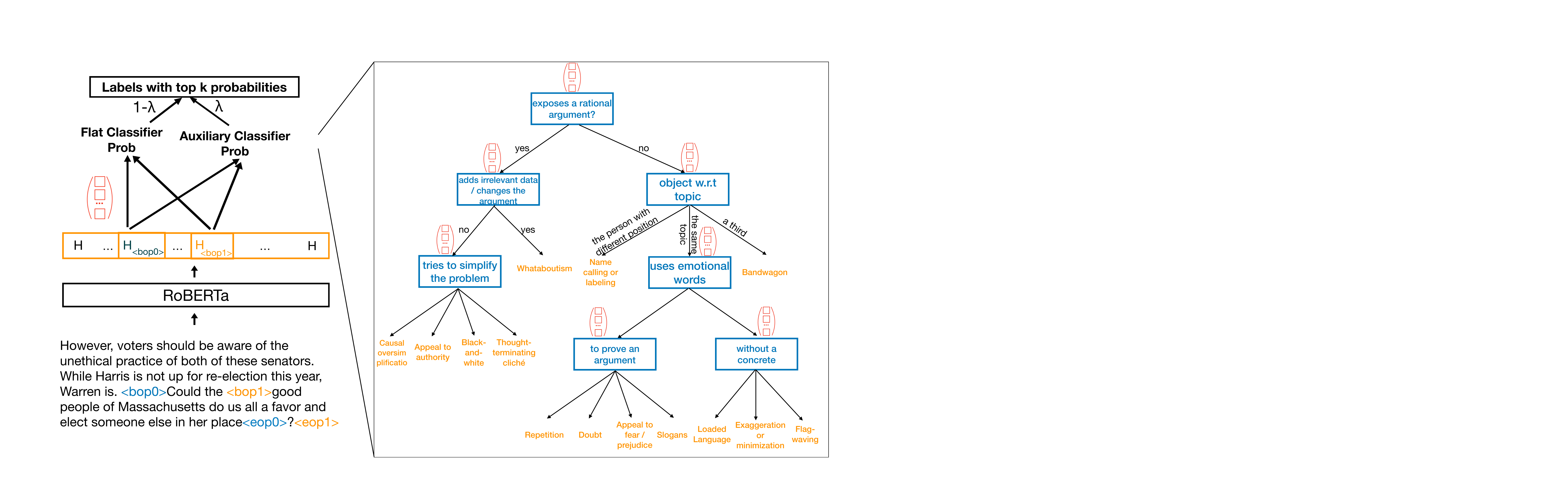}
\end{center}
\caption{Overview of the hier \mimlroberta.
Multiple spans within the same text (instances) are classified together. We add an auxiliary classifier to the model at each intermediate node. Decision tree used for annotations (right). }
\label{fig:aux_archi}
\end{figure*}

Besides, as a decision tree was used to guide annotations \citep{task},
we explore incorporating this hierarchical relationship among the labels into classifiers.
To do so, we add $7$ more auxiliary classifiers on top of the span representations from RoBERTa,
one for each intermediate node in the tree, and train these classifiers to predict
the path to a leaf node and hence the corresponding label (see Figure~\ref{fig:aux_archi}).
% Both the auxiliary classifiers and the original \emph{flat} $14$-way classifier are trained in a
% multi-task fashion, and their predictions are ensembled at test time.
We show that incorporating the label hierarchy in this manner improves both the single-instance and
the MIML versions of the RoBERTa model (referred to as hierarchical baseline and hierarchical $\text{MIML}_\text{RoBERTa}$ respectively).

% Hence, we apply hierarchical modeling to both the baseline and $\text{MIML}_\text{RoBERTa}$:
% in addition to the original linear classifier, we introduce 7 more classifiers, one at each parent node of the decision tree constructed out of the hierarchical diagram. These auxiliary classifiers also serve as useful regularizers to prevent overfitting the training data. A slight improvement is observed in both methods.

% 1. we decide to focus on sem eval. how the dataset looks like 
% 2. while others approach it as ..., MIML problem. cite MIML paper, show our similarities to theirs; one advantage of this method is model the interdependency among labels 
% 3. decision tree; hierarchical explicit 

% (TODO: http://kt.ijs.si/DragiKocev/repos/PR_Dimitrovski.pdf), medical terminologies(https://bmcbioinformatics.biomedcentral.com/articles/10.1186/1471-2105-13-161) and even social texts (https://dl.acm.org/doi/pdf/10.1145/2600428.2609595. 

% In this paper, we highlighted the benefits of employing hierarchical information among the labels t f
\section{Related Work}
The MIML framework was first introduced by \citet{miml} aiming for better representing complicated objects composed of multiple instances, e.g., an image with several bounding boxes each with its own label.
Since then, it has been applied to many tasks, such as relation extraction \cite{relation} and aspect-category
sentiment analysis \cite{acsa}.
The latter work uses a Bi-LSTM architecture which aggregates over the words in a
sentence (instances) to classify the sentiments of different aspects (labels).
% It was also applied in text classification previously in the proposed MIMLSVM and MIMLBoost \cite{miml}.
MIML has also been applied to BERT-based models in the biomedical text
analysis \citep{med3, med5}.
% and aspect-category
% sentiment analysis \cite{acsa}.
To our best knowledge, our work is the first one to apply it for propaganda technique classification.

Given the decision tree used for annotations (reproduced in Figure \ref{fig:labels_diagram} in Appendix), this task can also be viewed as a hierarchical text classification problem, with mandatory leaf node prediction and a tree-structured hierarchy \citep{silla2011survey}. Exploiting hierarchical information has been useful in significantly enhancing the performance of the system for medical image annotations \citep{med_image} and presents us with an opportunity to apply to the propaganda method detection. 
Similar to \citep{dumais2000hierarchical,weigend1999exploiting},
we use the multiplicative rule to combine probabilities along the different paths
in the hierarchy, leading to a distribution across the leaf nodes which can be combined
with the distribution predicted by the non-hierarchical baseline. Different approaches to hierarchical classification across multiple domains can be found in \citet{silla2011survey}.

% This task can be seen as a sequence labeling problem. BERT-based models have shown satisfactory performance in sequence labeling without the use of Conditional Random Fields given BERT's inherent attention to the surrounding context \citep{BERT}. 

% For easier integration with the baseline model which interpret the output as the probability over the label space, we use the multiplicative rule to combine probabilities down the path of different levels to produce the auxiliary probability, an idea similar to the one used in \citep{dumais2000hierarchical, weigend1999exploiting}.
% As far as we know, this is the first system that incorporates hierarchical information about the labels in propaganda technique classification.

%%%%%%%%%%%%%%%
%%%%%%%%%%%%%%

% Overview of the Hierarchical \mimlroberta system.
% Multiple spans within the same text (instances) are classified together.
% On the right we show the decision tree used for annotating instances
% with propaganda techniques. We add an auxiliary classifier to the model
% for each intermediate node in the tree.

\section{Methods\footnote{Code is available at \url{https://github.com/Dranoxgithub/propaganda-nlp-new}.}}
\label{sec:methods}
% In this section, starting from the single-instance baseline from the winning
% system at SemEval $2020$ (\S~\ref{sec:baseline}),
% we describe our approaches for MIML learning (\S~\ref{sec:miml})
% and incorporating the label hierarchy (\S~\ref{sec:hierarchical}).

\paragraph{Task Description.} Given an article $d$ with propagandistic spans 
$s_1, ..., s_m$ identified by their start and end indices,
% and the size of each of the label sets $|y_i| \forall i$,
the task is to identify the sets of techniques $y_1, ..., y_m$ where $y_i \subseteq Y \enskip \forall i \in \{1, ..., m\}$ and $Y$ represents the set of 14 techniques.
Following the shared task, we assume that the number of labels
$|y_i|$ for each span is known.
\subsection{Single-Instance Baseline}
\label{sec:baseline}
The baseline system from the winning ApplicaAI team \citep{applicaai} uses a
RoBERTa-based classifier and applies to each span separately.
The span is padded by special tokens \texttt{\textless{}bop\textgreater{}} and \texttt{\textless{}eop\textgreater{}} on either side.
A total context of $256$ tokens on both sides of the target span spanning multiple sentences are included in the input, unlike other systems \citep{aschern, hitachi, solomon} on the leaderboard which limit to the target sentence.
For classification, we use the \texttt{\textless{}bop\textgreater{}} representation output from RoBERTa
and pass it through a linear layer followed by softmax.
% Both the training and the inference is single-instance based, meaning one text fragment at a time.  
\citet{applicaai} further show marginal improvements by re-weighting the loss
for under-represented classes,
self-training over unlabeled data
and a Span-CLS approach which adds a second transformer network on top of
RoBERTa only over the tokens in the target span.
In our experiments, we did not use any of the improvements.
\subsection{\mimlroberta}
\label{sec:miml}
During the initial inspection of data, we observed high absolute pointwise mutual information between certain techniques. For example, in an article, if the \textit{slogans} technique appears, the \textit{flag-waving} technique usually follows.
% This makes intuitive sense as both techniques are tapping on strong emotions and that an article very likely follows the same writing style of emotional appeals.
This observation motivates us to predict the labels of \textit{all} spans in a text
simultaneously.
% the use of MIML to better model the interdependencies among the labels.

% (See Figure TODOQ in Appendix for an example). Here's a sample sentence and how it looks like before and after truncation at ``liar'':

% \textcolor{blue}{<bop0>}\textcolor{red}{<bop1>}You lied to me, didn’t you\textcolor{red}{<eop1>}?\\
    % \textcolor{orange}{<bop1>}You’re a liar\textcolor{orange}{<eop1>}, aren’t you\textcolor{blue}{<eop0>}?
% \textcolor{blue}{<bop0>}\textcolor{red}{<bop1>}You lied to me, didn’t you\textcolor{red}{<eop1>}?\\
    % \textcolor{orange}{<bop1>}You’re a \textcolor{orange}{<eop1>} \textcolor{blue}{<eop0>}
    
% Preprocessing details can be found in Appendix \ref{app:processing} and an example input is shown in Figure~\ref{fig:aux_archi}.

% During training, we take the representation of the span \texttt{\textless{}bop\textgreater{}}s and pass them through a linear layer
% to obtain logits for each of the $14$ techniques.

% we attach numbered \texttt{\textless{}bop\textgreater{}}s
% and \texttt{\textless{}eop\textgreater{}}s to every span and split article into widnows of size 512 tokens with a stride of 256 until there are no more labeled spans (details in Appendix~\ref{app:processing}).

During preprocessing, we pad the spans with pairs of \texttt{\textless{}bop0\textgreater{}}, \texttt{\textless{}eop0\textgreater{}}, \texttt{\textless{}bop1\textgreater{}}, \texttt{\textless{}eop1\textgreater{}}... to indicate the start and the end of a text fragment, even in cases where some spans are nested inside or overlapping with others.
After padding, every article is split into windows of size $512$ tokens with a stride of $256$ tokens until there are no more labeled spans.
Whenever there are spans that need to be truncated as we create a window, its corresponding \texttt{\textless{}eop\textgreater{}} is appended near the end of the window to ensure that the number of \texttt{\textless{}bop\textgreater{}}s and \texttt{\textless{}eop\textgreater{}}s match.
In the case where there is a specific nesting structure between any two text fragments, its nesting structure is respected in appending the \texttt{\textless{}eop\textgreater{}} to represent the differences in text fragments even after truncation. Also, since only around $1.8\%$ of the total annotations are spans with multiple labels \citep{task}, for ease of implementation, we use the rarer label for every text span during preprocessing and the later training.

Let $h_s$ denote the RoBERTa output
at the \texttt{\textless{}bop\textgreater{}} for a span $s$, then we compute the logits
for each of the labels as $p_\text{flat}(c) \propto \exp(h_s \cdot w_c)$,
for $c =0, \ldots, 13$. The model is trained using the cross-entropy loss: 
\begin{align}
    l_\text{flat} = \sum_{c = 1}^C  - \mathbbm{1}_{y=c} \log p_\text{flat}(c),
\end{align}
\noindent where $w_c$ is a weight vector and $y$ is the ground truth label of the span.
We compute $l_\text{flat}$ simultaneously for all spans in the
text and take an average.
% The loss is averaged across all spans in an example and then across all examples in a batch.
% the total number of spans in the text, $C$ is the number of classes
% and $y_s$ is the target label for the span $s$.
% $x$ is the input, $y$ is the target, $C$ is the number of classes, $S_n$ is the number of text fragments in the $n$-th element of the batch, and N refers to the batch size. 

During inference, we predict the number of labels requested for each span by selecting
the classes with the highest in $p_\text{flat}$. Since some spans can appear in multiple windows, the set of predictions from the window where the span is least truncated and has the most surrounding context is chosen. 
\subsection{Hierarchical $\text{MIML}_\text{RoBERTa}$}
\label{sec:hierarchical}

Due to the complex and subjective nature of the task, span annotations by \citet{task} were guided by a decision tree.
For example, annotators first consider whether the argument is rational, followed by whether emotional words are used, and so on.
% The annotation diagram and the constructed decision tree are in Figures \ref{fig:labels_diagram} and \ref{fig:tree} in Appendix.
This process induces a hierarchy among the labels,
% whereby certain
% labels are closely related to each other (and more likely to be confused
% with each other).
and to model such relationships, we use a simplified version of the decision tree
whose leaf nodes are the $14$ propaganda techniques (Figure~\ref{fig:aux_archi}, right).
We add an auxiliary loss to the training objective based
on hierarchical text classification which trains
local classifiers at each intermediate node \cite{silla2011survey}.
% Specifically,
% we add a local classifier for each intermediate node which,
% given a span,
% is trained to predict the outgoing edge which leads to the leaf
% node representing the label of that span \cite{silla2011survey}.
% Since spans with more than one label are rare, for ease of implementation of the auxiliary classifiers, we take the rarer label and ignore the others for a text fragment with multiple labels.
% Because of the need to have tight control over the number of predicted labels as demanded by the shared task, we use the local classifier per parent node hierarchical approach, meaning one classifier at every node of the decision tree, instead of a local classifier per node which uses a binary classifier at every possible label node, or local classifier per level which trains a single multi-class classifier at each level of the hierarchy \citep{silla2011survey}. 

Let $K$ be the number of intermediate nodes and hence the number of auxiliary classifiers,
and let $C_1, C_2, \ldots, C_K$ denote the number of outgoing edges or the number of labels
for each classifier.
Then, given the RoBERTa representation for a span $h_s$, we compute the probability
of following an edge $i$ from an intermediate node $k$ as:
\begin{equation}
    p_{k}(i) \propto \exp(h_s \cdot w_{k, i}), \quad \forall i = 0, \ldots, C_k
\end{equation}
\noindent where $w_{k, i}$ is a weight vector and the probabilities are normalized
across the $C_k$ labels for each classifier.
Given a leaf node label $c$, we denote the path of classifiers
and edges from the root to it as $I_c = \{(k_1, i_1), (k_2, i_2), \ldots\}$,
where each tuple in the set denotes a pair of classifier and its corresponding label along
the path.
Then we can compute the overall probability of selecting as:
\begin{equation}
    p_\text{aux}(c) = \prod_{(k, i)\in I_c} p_k(i).
\end{equation}

Note that $p_\text{aux}$ forms a valid distribution over the leaf nodes since each probability along the path is normalized.

During training, we compute an auxiliary loss for each span
which minimizes the negative log-likelihood of selecting the
edges along the path to its correct label $y$:
\begin{equation}
    l_\text{aux} = \frac{1}{|I_y|}\sum_{(k, i) \in I_y} \sum_{c=1}^{C_k} -\mathbbm{1}_{i=c} \log p_k(i).
\end{equation}

$\lambda_\text{training}$ and $\lambda_\text{eval}$ are both hyperparameters. The overall training loss is a combination of the flat loss described in the previous section and
the auxiliary loss above: $l_\text{ovr} =  (1 - \lambda_\text{training}) \cdot l_\text{flat} + \lambda_\text{training} \cdot l_\text{aux}.$
% \begin{equation}
%     l_\text{ovr} = \lambda_\text{training} * l_\text{flat} + (1 - \lambda_\text{training}) * l_\text{aux}.
% \end{equation}
% \vspace{-0.1cm}
During inference, we combine the predicted probabilities over the labels from both the
flat classifier and auxiliary classifiers: $p_\text{ovr}(c) = (1 - \lambda_\text{eval}) \cdot p_\text{flat}(c) + \lambda_\text{eval} \cdot p_\text{aux}(c).$

\section{Experiments}
\subsection{Dataset}

The original training and development set respectively contains $357$, $74$ articles, and thus correspondingly $6128$, $1063$ data points.  
Following \citet{applicaai}, we also evaluate using six-fold cross-validation where the
folds are created by mixing the original training and development sets. As a result each fold
roughly consists of  $6000$ training and $1200$ evaluation data points. Below we report the average and standard deviation of the metrics across six folds.
We also submit our best models, trained on the original training set, to the leaderboard for evaluation on the official test set.
\subsection{Training \& Evaluation}
\begin{figure}
\centering
\small
\begin{center}
\includegraphics[width=8cm]{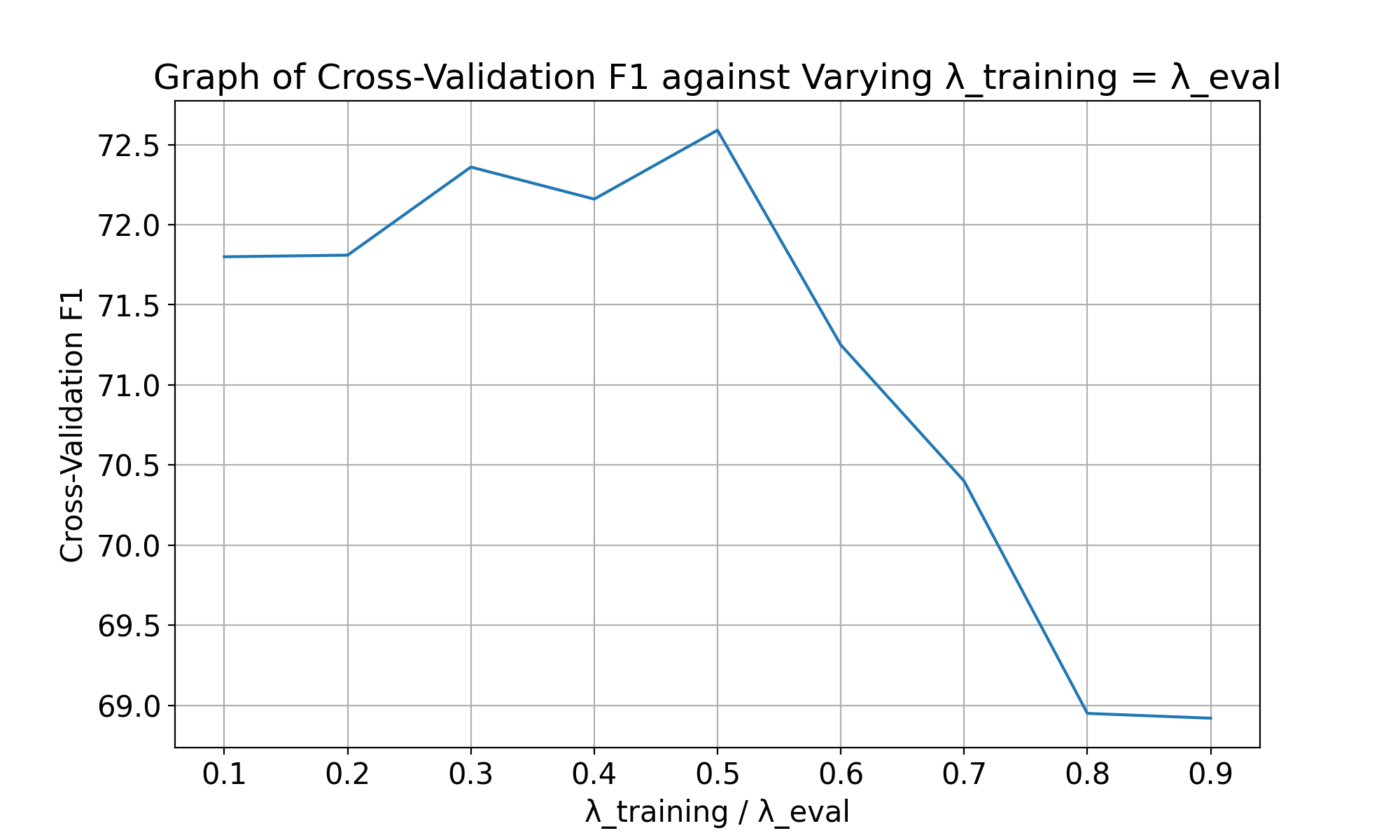}
\end{center}
\caption{Cross-validation F1 against varying lambda values where $\lambda_\text{training} = \lambda_\text{eval}$. The performance is the best at $\lambda_\text{training} = \lambda_\text{eval} = 0.5$.}
\label{fig:varying_lambda}
\end{figure}

We tune both $\lambda_\text{training}$ between $0$ and $1$ with a step of $0.1$. For every value of $\lambda_\text{training}$, we also tune $\lambda_\text{eval}$ at 0, 1 and $\lambda_\text{training}$.The model's performance peaks at $\lambda_\text{training} = \lambda_\text{eval} = 0.5$ (Figure~\ref{fig:varying_lambda}) and all the results reported use this set of values. 
Hyperparameter details are included in Appendix~\ref{app:hyperparams}.

% \begin{figure}
% \centering
% \includegraphics[width=8cm]{figures/truncation.pdf}
% \caption[Illustration of how the truncation respects the nesting structure in the preprocessing of the chunk method]{Illustration of how the truncation respects the nesting structure. The difference in text fragments (above and below the dotted line) is exemplified in different nesting structures before truncation. The difference is also preserved after truncation by the differences in the ordering of \textless{}eop0\textgreater{} and \textless{}eop1\textgreater{}. } \label{fig:truncation_chunk}
% \end{figure}

% To establish comparable standards with the results reported from the ApplicaAI team, we use the same six folds which are created from the original training and the development sets. All the cross validation results reported below is referring to the average and standard deviation of the results from the six folds. 
The scorer script provided by the shared task organizers evaluates a micro-averaged F1 score,
taking the best match between the predictions and ground truth labels when a span has multiple labels \citep{task}.
We use this script for evaluation in all our experiments.
% We evaluate the results using the script in the cross validation setting and on the original test set. 
Additionally, we also use a \textit{Tree-F1} score from the hierarchical text
classification literature which measures the overlap between the paths from
the root node to the ground truth and predicted nodes \cite{tree-f1}
(details in Appendix~\ref{app:tree-f1}).
In particular, we are interested in seeing if models with hierarchical information
make mistakes closer to the ground truth in the tree.
% Given the label hierarchy, one might also expect that a model should make mistakes closer
% to the ground truth labels in the tree.
% To measure this, we report a \emph{Tree-F1} score from the hierarchical
% text classification literature \cite{tree-f1}. For text span with multiple labels, we also take the best match.
\section{Results and Discussion}
% \begin{table*}
% \centering
% \begin{tabular}{lll}
% \hline
% \textbf{Cross Validation Micro-F1}   & \textbf{Non-hierarchical} & \textbf{Hierarchical}\\
% \hline
% Baseline  & $70.12 \pm 2.06$ & $70.49 \pm 2.01$\\
% $\text{MIML}_\text{RoBERTa}$ & $72.10 \pm 0.83$ & $72.59 \pm 1.02$\\
% % $72.29 \pm 0.95$ 
% \hline
% \end{tabular}
% \caption{\label{tab:cross_valid_F1}
% Micro-F1 of various methods in the cross validation setting. The best performing model out of all different $\lambda_\text{training}$ and $\lambda_\text{eval}$ is reported.}
% \end{table*}

% \begin{table*}
% \centering
% \begin{tabular}{lll}
% \hline
% \multicolumn{2}{l}{\textbf{Method}}   & \textbf{Cross validation F1} \\
% \hline
% \multirow{2}{*}{Baseline} & non-hierarchical & $70.12 \pm 2.06$ \\
% & hierarchical & $70.49 \pm 2.01$\\
% \hline
% \multirow{3}{*}{$\text{MIML}_\text{RoBERTa}$} & non-hierarchical & $72.10 \pm 0.83$\\
% & hierarchical & $72.59 \pm 1.02$\\
% & random & $69.54 \pm 1.33$\\
% \hline
% \end{tabular}
% \caption{\label{tab:results}
% Performance of various methods. The best performing model out of all different $\lambda_\text{training}$ and $\lambda_\text{eval}$ is reported.
% \bd{We should report the test set results in the separate table where we
% show more systems on the leaderboard.
% Also, can we get confidence intervals for the Tree-F1 scores as well?}
% \bd{We should move the tree-F1 scores to a separate table too -- they are
% distracting from the main results here.
% Also add the random hierarchy baseline here?}
% }
% \end{table*}

\begin{table}[!htbp]
\small
\centering
\begin{tabular}{llc}
\toprule
\multicolumn{2}{l}{\textbf{Method}}   & \textbf{Cross validation F1(\%)} \\
\midrule
\multirow{2}{*}{Baseline} & non-hier & $70.12 \pm 2.06$ \\
& hier & $70.49 \pm 2.01$\\
\midrule
\multirow{3}{*}{\mimlroberta} & non-hier & $72.10 \pm 0.83$\\
& random & $69.54 \pm 1.33$\\
& hier & $\mathbf{72.59 \pm 1.02}$\\
\bottomrule
\end{tabular}
\caption{\label{tab:cross_f1}
Mean and standard deviation of the F1 score across $6$ folds
for the single-instance baseline and \mimlroberta,
with and without hierarchical loss.
\textit{random} refers to an ablation where we randomly shuffle the nodes
in the hierarchy.
% We report the best configuration of $\lambda_\text{training}$ and $\lambda_\text{eval}$.
}
\end{table}

% \begin{table}
% \centering
% \begin{tabular}{llc}
% \hline
% \multicolumn{2}{l}{\textbf{Method}} & \textbf{Tree-F1} \\
% \hline
% \multirow{2}{*}{Baseline} & non-hier & $51.19$ \\
% & hier & $51.71$\\
% \hline
% \multirow{3}{*}{$\text{MIML}_\text{RoBERTa}$} & non-hier & $51.33$\\
% & hier & $51.64$\\
% \hline
% \end{tabular}
% \caption{\label{tab:tree_F1}
% Tree-F1 for incorrect data points of various methods. }
% \end{table}

Table~\ref{tab:cross_f1} shows the micro-averaged F1 scores of the different models discussed
in Section~\ref{sec:methods}.
We see a significant improvement when using the MIML framework: in cross validation, \mimlroberta~ has a micro-F1 of $72.10\%$, an absolute improvement of $1.98\%$ compared to the baseline single-instance model ($70.12\%$).
This improvement holds whether we use the hierarchical loss or not --
the hierarchical \mimlroberta has a $2.1\%$ improvement over the hierarchical baseline ($70.49\%$).
% This indicates that the MIML framework is useful to help the systems learn the interdependencies between labels and hence give more accurate predictions. 

\begin{table}[!htbp]
\small
\centering
\begin{tabular}{@{}llcc@{}}
\toprule
\multirow{2}{*}{\textbf{Method}} & \textbf{} & \multicolumn{2}{c}{\textbf{Tree-F1(\%)}}  \\ \cmidrule(l){3-4}
                                 & \textbf{} & \textbf{(Incorrect)} & \textbf{(All)} \\ \midrule
\multirow{2}{*}{Baseline}        & non-hier  & 51.19                & 85.44          \\
                                 & hier      & 51.71                & 85.78          \\ \midrule
\multirow{2}{*}{\mimlroberta}            & non-hier  & 51.33                & 86.53          \\
                                 & hier      & 51.64                & 86.76          \\ \bottomrule
\end{tabular}
\caption{\label{tab:tree_F1}
Tree-F1 scores for incorrect predictions and all predictions across
different models.
Incorporating $\text{loss}_\text{aux}$ improves the Tree-F1 score in both cases.
}
\end{table}

We also observe small but consistent improvements when training and predicting with
hierarchical information in the form of auxiliary classifiers:
% Also, there is a further small improvement when integrating hierarchical information: after the integration, for cross validation micro-F1,
the baseline model improves by $0.37\%$, while \mimlroberta
improves by $0.49\%$.
Table~\ref{tab:tree_F1} shows the Tree-F1 scores over the full validation splits,
as well as only for the incorrect predictions from the various models.
Again, we observe a small improvement due to incorporating the hierarchical information --
specifically,
% Similar improvements can be found in the average Tree-F1 for all data points and the average Tree-F1 for incorrectly classified data points. This is in line with our expectation. With the learning of the hierarchical information,
the incorrect predictions are now closer to the ground truth labels,
as evidenced by the higher Tree-F1 over the mistakes.
% predictionsand hence the tree-F1 for the incorrectly classifier data points should be higher.

To further confirm that these improvements are due to learning hierarchical information rather than any regularization effect from the additional auxiliary loss term, we also run the experiments with the labels on the decision tree randomly shuffled with the same lambda values. We obtain a micro-F1 of $69.54$ with $\lambda_\text{training} = \lambda_\text{eval} = 0.5$, which is significantly lower. Particularly, when $\lambda_\text{training} = 0.5$ and $\lambda_\text{eval} = 1$, i.e. inference is done using only the auxiliary classifiers, the micro-F1 in cross validation is an extremely low $5.25\%$, in contrast to a $72.33\%$ when not shuffled.
% This demonstrates that the auxiliary classifiers are indeed learning the hierarchical information among the labels and that the minor improvement is due to this learning instead of mere regularization effects.  (TODOQ: contradition to the intro??)

\begin{table}[!htbp]
\small
\centering
\begin{tabular}{llc}
\toprule
\multicolumn{2}{l}{\textbf{System}} & \textbf{Test F1}\\
\midrule
\multirow{3}{*}{Single} & \citet{singh-etal-2020-newssweeper} & $58.436$\\
& \citet{nopropaganda} & $59.832$\\
& non-Hier \mimlroberta~(Ours) & $62.179$\\
& Hier \mimlroberta~(Ours) & $62.793$\\
\midrule
\multirow{3}{*}{Ensemble} & \citet{hitachi} & $63.129$ \\
& \citet{aschern} & $63.296$ \\
& \citet{applicaai}  & $63.743$\\
\bottomrule
\end{tabular}
\caption{\label{tab:test_f1}
F1 score on the official test set of various systems on the leaderboard. See \url{https://propaganda.qcri.org/ptc/leaderboard.php}
}
\end{table}

% \begin{table}[!htbp]
% \small
% \centering
% \begin{tabular}{lc}
% \toprule
% \textbf{System} & \textbf{Test Set F1(\%)}\\
% \midrule
% Single Models\\
% \midrule
% \citet{singh-etal-2020-newssweeper} & $58.436$\\
% \citet{nopropaganda} & $59.832$\\
% Hier \mimlroberta~(Ours) & $62.793$\\
% \midrule
% Ensemble Models\\
% \midrule
% % \citet{solomon} & $59.386$ \\
% \citet{hitachi} & $63.129$ \\
% \citet{aschern} & $63.296$ \\
% \citet{applicaai}  & $63.743$\\
% \bottomrule
% \end{tabular}
% \caption{\label{tab:test_f1}
% F1 score on the official test set of various systems on the leaderboard. See \url{https://propaganda.qcri.org/ptc/leaderboard.php}
% }
% \end{table}
% \vspace{-0.3cm}

The micro-F1 results on the official test set are in Table~\ref{tab:test_f1},
where we see that the hierarchical \mimlroberta is the best performing single model.
Other than improving accuracy, \mimlroberta also reduces the training and evaluation time,
since it predicts multiple spans in a single forward pass through the RoBERTa model.
% This saving on time is because, with the preprocessing, the use of the chunk has 2059 windows, much fewer data points than the span method which has 7128 text fragments and hence 7128 data points.
One epoch of \mimlroberta (one evaluation every $25$ steps) takes $5.04$ minutes
compared to $15.50$ minutes by the single-instance baseline, a $68\%$ reduction.
% For every epoch which trains with an evaluation every 25 steps, as compared to the span method's 15.50 minutes, the chunk method only takes 5.04 minutes (a $67.48\%$ reduction) while the auxiliary classifier takes 5.23 minutes (a $66.26\%$ reduction). 

\section{Conclusion}
We propose two simple extensions to a RoBERTa-based model for propaganda technique classification,
which lead to notable improvements.
Our approach to incorporating hierarchical information about the labels into training could
also be useful for other tasks where the annotation procedure involves making a series
of decisions about instances.
Future work can also explore other methods to incorporate the hierarchical information,
e.g., via regularizing the label embeddings.

\section*{Limitations}
The use of auxiliary classifiers at every node of the decision tree is not feasible when the hierarchical tree is huge, such as the large hierarchical terminologies for medical literature indexing \citep{gasco2021overview}. 

Besides, in Table~\ref{tab:cross_f1}, even though the integration of the hierarchical information shows a consistent improvement in both the baseline and \mimlroberta models, these improvements are still within one standard deviation of micro-F1. 

Lastly, it is worth noting that we do not focus on large language models since our approach is to explore improvements on a published state-of-the-art model. While they might improve accuracy, a careful exploration of those on a new task is beyond the scope.

\section*{Ethics Statement}
Propaganda detection is a sensitive topic and any practical application of the model needs to be carefully orchestrated. Both false positives and false negatives of the model can have harmful impacts. Moreover, there might be certain biases in the training data, and consequently this leads to systematic issues in the model, such as a higher tendency to mislabel certain kinds of text. 

Furthermore, this paper follows the same definition as SemEval $2020$ Task $11$ \citep{task} whereas there might a broader debate on the definition of propaganda. 

Lastly, we also acknowledge the concern that a perfect classifier for propaganda text can be used to train a language model that generates propaganda which in turn evades the classifier's detection. 

\bibliography{emnlp2021}
\bibliographystyle{acl_natbib}

\begin{table*}
\small
\centering
\begin{tabular}{c c c c c}
\toprule
Hyperparameter & Baseline & Hier Baseline & $\text{MIML}_\text{RoBERTa}$ & Hier $\text{MIML}_\text{RoBERTa}$\\
\midrule
Dropout & 0 & 0 & 0.1 & 0.1\\ 
 Learning Rate & 2e-5 & 2e-5 &1e-5 & 1e-5\\   
 Weight decay & 0.01 & 0.01 & 0.1 & 0.1\\  
 \midrule
 Loss & BCE & BCE & CE & CE\\
  Batch size  & 16 & 16 & 8 & 8\\  
  \midrule
Context Size & 512 & 512 & 512 & 512 \\
 Number of epochs & 20 & 20 & 20 & 20\\ 
 Optimizer & AdamW &  AdamW& AdamW & AdamW \\  
 \bottomrule
\end{tabular}
\caption{Optimizers and hyperparameters for all different methods.} \label{tab:param}
\end{table*}

\newpage
\appendix

% \section{\mimlroberta Preprocessing \label{app:processing}}
% Since only around $1.8\%$ of the total annotations are spans with multiple labels \citep{task}, for the ease of implementation, we use the rarer label for every text span during training.

% During preprocessing, we pad the spans with pairs of \texttt{\textless{}bop0\textgreater{}}, \texttt{\textless{}eop0\textgreater{}}, \texttt{\textless{}bop1\textgreater{}}, \texttt{\textless{}eop1\textgreater{}} to indicate the start and the end of a text fragment, even in cases where some spans are nested inside or overlapping with others.
% After padding, every article is split into windows of size $512$ tokens with a stride of $256$ tokens until there are no more labeled spans.
% Whenever there are spans that need to be truncated as we create a window, its corresponding \texttt{\textless{}eop\textgreater{}} is appended near the end of the window to ensure that the number of \texttt{\textless{}bop\textgreater{}}s and \texttt{\textless{}eop\textgreater{}}s match.
% In the case where there is a specific nesting structure between any two text fragments, its nesting structure is respected in appending the \texttt{\textless{}eop\textgreater{}} to represent the differences in text fragments even after truncation.

\begin{figure}
\centering
\includegraphics[width=8cm]{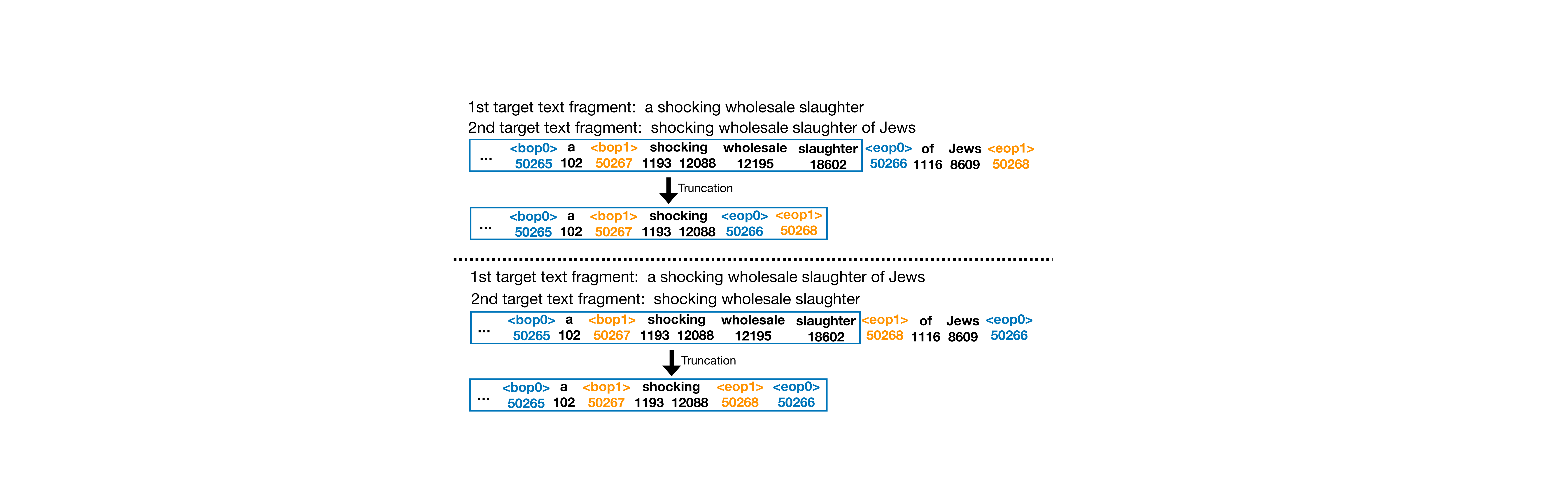}
\caption[Illustration of how the truncation respects the nesting structure in the preprocessing of the chunk method]{Illustration of how the truncation respects the nesting structure. The difference in text fragments (above and below the dotted line) is exemplified in different nesting structures before truncation. The difference is also preserved after truncation by the differences in the ordering of \textless{}eop0\textgreater{} and \textless{}eop1\textgreater{}. } \label{fig:truncation_chunk}
\end{figure}

\section{Techniques}
\begin{table}[!htbp]
\centering
\begin{tabular}{ll}
\toprule
\textbf{Technique} & \textbf{Freq} \\ 
\midrule
 Loaded Language & 2448  \\ 
 Name calling or labeling & 1241 \\  
 Repetition & 766  \\   
 Exaggeration or minimization & 534\\   
 Doubt & 559 \\   
 Appeal to fear/prejudice & 338 \\  
 Flag-waving & 316 \\
 Causal oversimplification & 227 \\ 
 Slogans & 169 \\ 
 Appeal to authority & 158 \\
 Black-and-white fallacy, dictatorship & 129 \\
Thought-terminating cliché & 93 \\ 
Whataboutism, straw man, red herring & 136 \\
Bandwagon, reductio ad hilterum & 77 \\ 
\bottomrule
\end{tabular}
\caption{\label{tab:technique_freq} Summary of techniques and their frequency in the data. The definitions of the techniques are found in \url{https://propaganda.qcri.org/annotations/definitions.html}}
\end{table}

\begin{figure*}[htp]
\begin{center}
\includegraphics[width=\textwidth]{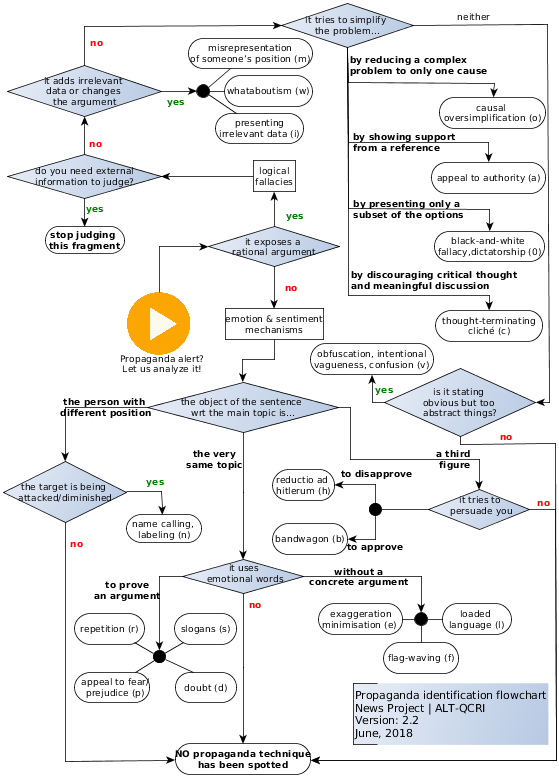}
\end{center}
\caption[The hierarchical diagram for annotators]{The hierarchical diagram to guide the annotation of propaganda techniques \citep{task}.}
\label{fig:labels_diagram}
\end{figure*}

The hierarchical diagram to guide the annotation of propaganda techniques can be found in Figure~\ref{fig:labels_diagram}.

\section{Hyperparameters}
\label{app:hyperparams}

We use HuggingFace's library \citep{huggingface} and a single Nvidia RTX A6000 GPU. Both the baseline and the hierarchical baseline methods use a learning rate of $2e-5$, a dropout of $0$ and a batch size of $16$, while the $\text{MIML}_\text{RoBERTa}$ and its hierarchical version use a learning rate of $1e-5$, a dropout of $0.1$ and a batch size of $8$. All the models are trained for 20 epochs. More details can be found in Table~\ref{tab:param}. 

\section{Tree-F1 Metric}
\label{app:tree-f1}

Let $Y$ be the ground truth label and $Y'$ be the prediction, and let $K$ be
the lowest common ancestor between the two. Then the Tree-F1 is given by:
% To assess the incorporation of the hierarchical information among labels, we use set-based evaluation measures (further referred to as tree F1)\citep{tree-f1}: given a ground truth label $Y$ and a predicted label $Y'$, let their lowest common ancestor be $K$, 
% then
$$\text{Precision}_\text{Tree} = L_K / L_{Y'}$$
$$\text{Recall}_\text{Tree} = L_K / L_Y$$ 
$$\text{Tree-F1} = \frac{2 * \text{Precision}_\text{Tree} * \text{Recall}_\text{Tree}}{\text{Precision}_\text{Tree} + \text{Recall}_\text{Tree}}$$

where $L_M$ refers to the number of nodes tracing from the root node down to node $M$.

\end{document}